\documentclass{article}

\usepackage{microtype}
\usepackage{graphicx}
\usepackage{subcaption}
\usepackage{booktabs} 
\usepackage{multirow}

\usepackage{amsmath,amsfonts,bm}

\def\figref#1{Fig.~\ref{#1}}
\def\Figref#1{Fig.~\ref{#1}}

\def\eqref#1{Eq.~\ref{#1}}

\def\1{\bm{1}}

\DeclareMathAlphabet{\mathsfit}{\encodingdefault}{\sfdefault}{m}{sl}
\SetMathAlphabet{\mathsfit}{bold}{\encodingdefault}{\sfdefault}{bx}{n}

\def\gN{{\mathcal{N}}}

\newcommand{\E}{\mathbb{E}}

\usepackage{hyperref}
\usepackage{lipsum}

\usepackage[accepted]{icml2024}

\usepackage{amsmath}
\usepackage{amssymb}
\usepackage{mathtools}
\usepackage{amsthm}
\usepackage{mathrsfs}

\usepackage{pifont}
\usepackage[capitalize,noabbrev]{cleveref}

\theoremstyle{plain}

\theoremstyle{definition}

\theoremstyle{remark}

\usepackage[textsize=tiny]{todonotes}

\icmltitlerunning{Learning to Continually Learn with the Bayesian Principle}

\begin{document}

\twocolumn[
\icmltitle{Learning to Continually Learn with the Bayesian Principle}



\icmlsetsymbol{equal}{*}

\begin{icmlauthorlist}
\icmlauthor{Soochan Lee}{snu}
\icmlauthor{Hyeonseong Jeon}{snu}
\icmlauthor{Jaehyeon Son}{snu}
\icmlauthor{Gunhee Kim}{snu}
\end{icmlauthorlist}

\icmlaffiliation{snu}{Department of Computer Science and Engineering, Seoul National University, Seoul, Republic of Korea}

\icmlcorrespondingauthor{Gunhee Kim}{gunhee@snu.ac.kr}

\icmlkeywords{Machine Learning, ICML}

\vskip 0.3in
]



\printAffiliationsAndNotice{}  

\begin{abstract}
In the present era of deep learning, continual learning research is mainly focused on mitigating forgetting when training a neural network with stochastic gradient descent on a non-stationary stream of data.
On the other hand, in the more classical literature of statistical machine learning, many models have sequential Bayesian update rules that yield the same learning outcome as the batch training, i.e., they are completely immune to catastrophic forgetting.
However, they are often overly simple to model complex real-world data.
In this work, we adopt the meta-learning paradigm to combine the strong representational power of neural networks and simple statistical models' robustness to forgetting.
In our novel meta-continual learning framework, continual learning takes place only in statistical models via ideal sequential Bayesian update rules, while neural networks are meta-learned to bridge the raw data and the statistical models.
Since the neural networks remain fixed during continual learning, they are protected from catastrophic forgetting.
This approach not only achieves significantly improved performance but also exhibits excellent scalability.
Since our approach is domain-agnostic and model-agnostic, it can be applied to a wide range of problems and easily integrated with existing model architectures.
\end{abstract}

\section{Introduction}
\label{sec:intro}

Continual learning (CL), the process of acquiring new knowledge or skills without forgetting existing ones, is an essential ability of intelligent agents.
Despite recent advances in deep learning, CL remains a significant challenge.
\citet{Knoblauch2020OptimalCL} rigorously prove that, in general, CL is an NP-hard problem.
This implies that building a universal CL algorithm is impossible as long as P$\neq$NP.
To effectively tackle CL, one should first narrow down a domain and design a CL algorithm tailored to leverage a domain-specific structure.
Even humans possess specialized CL abilities for specific tasks, such as learning new faces, which may not be as effective for other tasks, such as memorizing random digits.
This specialization results from the evolutionary process that has optimized our CL abilities for survival and reproduction.

From this perspective, meta-continual learning (MCL) emerges as a highly promising avenue of research.
Rather than manually crafting CL algorithms based solely on human knowledge, MCL aims to meta-learn the CL ability in a data-driven manner -- \emph{learning to continually learn}.
Thus, we can design a general MCL algorithm and feed domain-specific data to obtain a specialized CL algorithm.
MCL can be more advantageous in many practical scenarios, as it can utilize a large-scale dataset to improve the CL ability before deploying a CL agent, instead of learning from scratch.

MCL follows the bi-level optimization scheme of meta-learning: in the inner loop, a model is continually trained by a CL algorithm, while in the outer loop, the CL algorithm is optimized across multiple CL episodes.
Although stochastic gradient descent (SGD) has been the primary learning mechanism in deep learning, this bi-level scheme offers the flexibility to combine neural networks with fundamentally different learning mechanisms.
Specifically, we can meta-train neural networks with SGD only in the outer loop and adopt another update rule for CL in the inner loop.

In this context, the sequential Bayesian update stands out as the most promising candidate, providing an ideal framework for updating a knowledge state.
While there have been a significant number of CL approaches inspired by the Bayesian updates of the posterior of neural network parameters \citep{EWC, SI, Chaudhry18Riemannian, VCL, Farquahr19Bayesian}, they require various approximations to ensure computational tractability, which sets them apart from the ideal Bayesian update.
On the other hand, we bring the Fisher-Darmois-Koopman-Pitman theorem \citep{Fisher1934, Darmois1935, Koopman1936, Pitman1936} into the scope to point out that the exponential family is the only family of distributions that are capable of efficient and lossless sequential Bayesian update (more precise description in \S\ref{sec:bg:exp}).
Instead of dealing with the intractable posterior of complex neural networks, we consider the sequential Bayesian inference of simple statistical models that inherently come with an exponential family posterior, yielding a result identical to batch inference.
While these models are immune to catastrophic forgetting by design, they are often too simple for modeling complex, high-dimensional data.
Fortunately, the MCL setting allows meta-training neural networks that can work as bridges between the real world and the statistical models.

We distill this idea of combining simple statistical models and meta-learned neural networks into a general MCL framework named \emph{Sequential Bayesian Meta-Continual Learning (SB-MCL)}.
Since SB-MCL is domain-agnostic and model-agnostic, it can be applied to a wide range of problem domains and integrated with existing model architectures with minimal modifications.
SB-MCL encompasses several prior works \citep{GeMCL, PN, ALPaCA} as special cases and supports both supervised and unsupervised learning.
In our extensive experiments on a wide range of benchmarks, SB-MCL achieves remarkable performance while using substantially less resources.
Code is available at \url{https://github.com/soochan-lee/SB-MCL}.

\section{Background}
\label{sec:bg}

\newcommand{\xtest}{\tilde x}
\newcommand{\ytest}{\tilde y}
\newcommand{\xtestn}[1]{\xtest_{#1}}
\newcommand{\ex}{(x, y)}
\newcommand{\exn}[1]{(x_{#1}, y_{#1})}
\newcommand{\exnk}[2]{(x_{#1}^{#2}, y_{#1}^{#2})}
\newcommand{\exntest}[1]{(\tilde x_{#1}, \tilde y_{#1})}
\newcommand{\dist}{F}
\newcommand{\trainset}{\mathcal{D}}
\newcommand{\trainsetdef}{\{\exn{n}\}_{n=1}^N}
\newcommand{\trainstream}{{\mathcal{D}}}
\newcommand{\trainstreamdef}{(\exn{t})_{t=1}^T}
\newcommand{\testset}{\mathcal{E}}
\newcommand{\testsetdef}{\{\exntest{n}\}_{n = 1}^{N}}
\newcommand{\epi}{(\trainset, \testset)}
\newcommand{\epim}[1]{(\trainset^{#1}, \testset^{#1})}
\newcommand{\clepi}{(\trainstream, \testset)}
\newcommand{\clepim}[1]{(\trainstream^{#1}, \testset^{#1})}

\newcommand{\kldiv}{D_{\mathrm{KL}}}

\subsection{Meta-Continual Learning}
We describe the problem setting of MCL.
We denote an example $\ex$ where $x$ is an input variable, and $y$ is a target variable, assuming a supervised setting by default.
For unsupervised learning settings, one can replace $\ex$ with $x$.
A CL episode $\epi$ consists of a training stream $\trainstream = \trainstreamdef$ and a test set $\testset = \testsetdef$.
The training stream is an ordered sequence of length $T$, and its examples can be accessed sequentially and cannot be accessed more than once.
It is assumed to be non-stationary and typically constructed as a concatenation of $K$ distinct \emph{task} streams.
Naively training a neural network on such a non-stationary stream with SGD results in catastrophic forgetting of the knowledge from the previous part of the stream.
The test set consists of examples of the tasks appearing in the training stream, such that the model needs to retain knowledge of all the tasks to obtain a high score in the test set.

In MCL, multiple CL episodes are split into a meta-training set $\mathcal D = \{(\trainstream^i, \testset^i)\}_i$ and a meta-test set $\mathcal E = \{(\trainstream^j, \testset^j)\}_j$.
During the meta-training phase, a CL algorithm is optimized across multiple episodes in $\mathcal D$ to produce a competent model from a training stream.
The algorithm's CL capability is then measured with $\mathcal E$.
Note that $\mathcal D$ and $\mathcal E$ typically do not share any underlying tasks since the meta-test set aims to measure the learning capability, not the knowledge of specific tasks that appear during meta-training.
Note that MCL should not be confused with other specialized settings that combine meta-learning and CL (\citealp{Finn19Online, RiemerCALRTT19, Jerfel19Reconciling, La-MAML}; to name a few).
They have different assumptions and objectives that are not compatible with MCL.

\subsection{Sequential Bayesian Update of Exponential Family Posterior}
\label{sec:bg:exp}
The Bayes rule offers a principled way to update knowledge incrementally by using the posterior at the previous time step as the prior for the current time step, i.e., $p(z | x_{1:t}) \propto p(x_t | z) p(z | x_{1:t-1})$ \citep{PRML, MurphyML1}.
Therefore, the Bayesian perspective has been widely adopted in CL research \citep{EWC, SI, Chaudhry18Riemannian, VCL, Farquahr19Bayesian}.
However, prior works have focused on sequentially updating the posterior of neural network parameters, which are generally intractable to compute.
Therefore, they must rely on various approximations, resulting in a wide gap between the ideal Bayesian update and reality.

Then, what kind of models are suitable for efficient sequential Bayesian updates?
According to the Fisher-Darmois-Koopman-Pitman theorem \citep{Fisher1934, Darmois1935, Koopman1936, Pitman1936}, \emph{the exponential family is the only family of distributions where the dimension of the sufficient statistic remains fixed, regardless of the number of examples}.
Sufficient statistics are the minimal statistics that capture all the information in the data about the parameter of interest.
Therefore, if the dimension of the sufficient statistic remains fixed, we can store all the necessary information in a fixed-size memory system.
This theorem has significant implications for CL; if the model's posterior is not a member of the exponential family (as in the case of neural networks) and does not have a large enough memory system to store the ever-growing sufficient statistics, forgetting becomes inevitable.
From this perspective, employing a replay buffer \citep{GEM, TinyMemory} is an approach that aids in partially preserving sufficient statistics.

On the flip side, the theorem suggests an alternative approach; by embracing an exponential family distribution, we can store sufficient statistics within a fixed dimension, enabling efficient sequential Bayesian updates without any compromises.
Although the exponential family's expressivity is limited, this challenge can be effectively addressed in MCL settings by meta-learning neural networks to reconcile the real-world data and the exponential family.

\section{Our Approach: SB-MCL}
\label{sec:method}

\newcommand{\truep}{p_\theta}  
\newcommand{\variq}{q_\phi}
\newcommand{\zmap}{z_\mathrm{MAP}}

\begin{figure}
    \centering
    \begin{subfigure}[b]{0.49\linewidth} 
        \centering
        \includegraphics[width=\linewidth, trim={0 539pt 263pt 45pt}, clip]{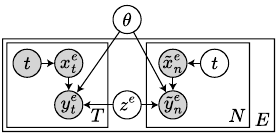}
        \caption{Supervised MCL.}
        \label{fig:gm:sup}
    \end{subfigure}
    \begin{subfigure}[b]{0.49\linewidth} 
        \centering
        \includegraphics[width=\linewidth, trim={0 605pt 263pt 0pt}, clip]{figures/main.pdf}
        \caption{Unsupervised MCL.}
        \label{fig:gm:unsup}
    \end{subfigure}
    \caption{
        Graphical models of MCL.
        For each episode $e$, training examples $(x_t^e, y_t^e)$ (or just $x_t^e$) are produced conditioned on the time step $t$ and the episode-wise latent variable $z^e$.
    }
    \label{fig:gm}
\end{figure}

\begin{figure}[t]
    \centering
    \includegraphics[width=\linewidth, trim={142pt 541pt 0 0}, clip]{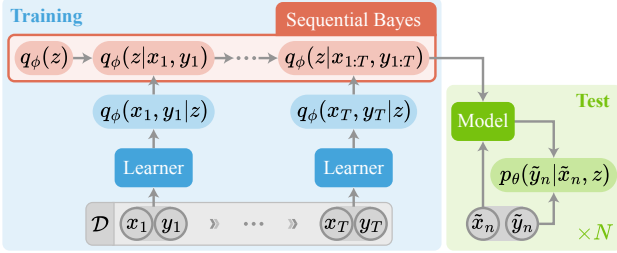}
    \caption{
        Schematic diagram of our SB-MCL in a single supervised CL episode.
        In SB-MCL, CL is formulated as the sequential Bayesian update of an exponential family posterior $\variq(z|x_{1:t}, y_{1:t})$.
        The meta-learned neural networks (the learner and the model) remain fixed during CL to protect themselves from catastrophic forgetting.
    }
    \label{fig:method}
\end{figure}

\subsection{The Meta-Learning Objective}
\label{sec:method:objective}

\Figref{fig:gm} shows the graphical models of our MCL settings.
In both supervised and unsupervised settings, there are $E$ CL episodes.
Each CL episode $e$ has a training stream $\trainstream^e$ of length $T$ and a test set $\testset^e$ of size $N$.
In supervised CL settings (\Figref{fig:gm:sup}), each example is a pair of input $x$ and target $y$, and the goal is to model the conditional probability $p(y|x)$.
In unsupervised settings (\Figref{fig:gm:unsup}), an example is simply $x$, and the goal is to model $p(x)$.
For each CL episode $e$, we assume an episode-specific latent variable $z^e$ that governs the entire episode.
The training stream's non-stationarity, a key characteristic of CL, is expressed by the time variable $t$ affecting the generation of $x$.
In practice, the training stream is often constructed by concatenating multiple \emph{task} streams, each of which is a stationary stream sampled from a distinct task distribution.
Note that $z^e$ is shared by all examples inside an episode regardless of the tasks they belong to.
Under this framework, the CL process is to sequentially refine the belief state of $z^e$.

The objective is to maximize the (conditional) log-likelihood of the test set $\testset$ after continually learning the training stream $\trainstream$ (superscript $e$ is now omited for brevity).
Assuming a model parameterized by $\theta$, this objective can be summarized as 
\begin{align}
\sum_{n=1}^N \log \truep(\ytest_n | \xtest_n, \trainstream)
= \sum_{n=1}^N \log \int_z \truep(\ytest_n | \xtest_n, z) \truep(z | \trainstream)
\nonumber
\end{align}
in supervised settings and as 
\begin{align}
    \sum_{n=1}^N \log \truep(\xtest_n | \trainstream)
    = \sum_{n=1}^N \log \int_z \truep(\xtest_n | z) \truep(z | \trainstream)
    \nonumber
\end{align}    
in unsupervised settings, where $\xtest_*$ and $\ytest_*$ are the test data in $\testset$. 
However, computing these objectives is generally intractable due to the integration over $z$.
For such cases, we introduce a variational distribution $\variq$ parameterized by $\phi$ and derive the variational lower bounds.
The bounds for the supervised and unsupervised cases are derived as

\begin{small}
\begin{align}
    &\log \truep(\ytest_{1:N} | \xtest_{1:N}, \trainstream) = \log \truep(\ytest_{1:N} | \xtest_{1:N}, x_{1:T}, y_{1:T}) \nonumber \\
    &\geq \E_{z \sim \variq(z | \trainstream)} \left[
        \sum_{n=1}^N \log \truep(\ytest_n | \xtest_n, z) + \sum_{t=1}^T \log \truep(y_t | x_t, z)
        \right] \notag \\
    &\quad - \kldiv \left( \variq(z | \trainstream) \, \middle\| \, \truep(z) \right) - \underbrace{\log \truep(\trainstream)}_{\mathrm{const.}}, \label{eq:elbo:sup}
\end{align}
\begin{align}
    &\log \truep(\xtest_{1:N} | \trainstream) = \log \truep(\xtest_{1:N} | x_{1:T}) \nonumber \\
    &\geq \E_{z \sim \variq(z | \trainstream)} \left[
        \sum_{n=1}^N \log \truep(\xtest_n | z) + \sum_{t=1}^T \log \truep(x_t | z)
        \right] \notag \\
    &\quad - \kldiv \left( \variq(z | \trainstream) \, \middle\| \, \truep(z) \right) - \underbrace{\log \truep(\trainstream)}_{\mathrm{const.}}. \label{eq:elbo:unsup}
\end{align}
\end{small}
For more details, please refer to Appendix \ref{app:derivation}.

\subsection{Continual Learning as Sequential Bayesian Update} 
\label{sec:method:cl}

In \eqref{eq:elbo:sup} and \ref{eq:elbo:unsup}, the CL process is abstracted inside the variational posterior $\variq(z | \trainstream)$, which is obtained through sequential Bayesian updates:
\begin{equation}
\begin{aligned}
    \variq(z | x_{1:t}, y_{1:t}) &\propto \variq(x_t, y_t | z) \variq(z | x_{1:t-1}, y_{1:t-1}), \\
    \variq(z | x_1, y_1) &\propto \variq(x_1, y_1 | z) \variq(z) \label{eq:bayes:sup}
\end{aligned}
\end{equation}
\begin{equation}
\begin{aligned}
    \variq(z | x_{1:t}) &\propto \variq(x_t | z) \variq(z | x_{1:t-1}), \\
    \variq(z | x_1) &\propto \variq(x_1 | z) \variq(z), \label{eq:bayes:unsup}
\end{aligned}
\end{equation}
where \eqref{eq:bayes:sup} and \ref{eq:bayes:unsup} are respectively for supervised and unsupervised CL.
In the following, we will consider only the supervised case to be concise, but the same logic can be extended to the unsupervised case.
As depicted in Fig.~\ref{fig:method}, The CL process initially starts with a variational prior $\variq(z)$.
And the \emph{learner}, a neural network component, produces $\variq(x_t, y_t | z)$ for each example $(x_t, y_t)$, which is subsequently integrated into the variational posterior $\variq(z | x_{1:t}, y_{1:t})$.\footnote{Both $\variq(x_{t}, y_{t} | z)$ and $\variq(z | x_{1:t}, y_{1:t})$ are used as functions of $z$ since $(x_{1:t}, y_{1:t})$ is given.}
The parameters of the prior and the learner constitute $\phi$.
As previously explained in \S\ref{sec:bg:exp}, the Fisher-Darmois-Koopman-Pitman theorem implies that only exponential family distributions can perform such updates without consistently increasing the memory and compute requirement proportional to the number of examples.
This property makes them ideal for our variational posterior.
Note that SB-MCL does not involve any gradient descent during CL; the learner performs only the forward passes to process the training examples for sequential Bayesian updates.

As an example of exponential family distributions, we describe the exact update rule for a factorized Gaussian posterior $\gN(z; \mu_t, \Lambda_t^{-1})$ where $\Lambda_t$ is diagonal.
\newcommand{\learnerz}{\hat{z}}
First, the variational prior is also defined as a factorized Gaussian: $\variq(z) = \gN(z; \mu_0, \Lambda_0^{-1})$.
For $\variq(x_t, y_t | z)$, the learner outputs $\learnerz_t$ and $P_t$ for each $(x_t, y_t)$, where $P_t$ is a diagonal matrix.
We consider $\learnerz_t$ as a noisy observation of $z$ with a Gaussian noise of precision $P_t$, i.e., $\variq(x_t, y_t | z) = \gN(\learnerz_t; z, P_t^{-1})$ \citep{Volpp21BayesContext}.
This allows an efficient sequential update rule for the variational posterior \citep{PRML}:
\begin{align}
    \Lambda_t = \Lambda_{t-1} + P_t,\hspace{3mm} \mu_t = \Lambda_t^{-1} \left( \Lambda_{t-1} \mu_{t-1} + P_t \learnerz_t \right). \label{eq:post:seq}
\end{align}

After training, the posterior $\variq(z | \trainstream) = \variq(z | x_{1:T}, y_{1:T})$ is passed on to the test phase.
During testing, the model produces outputs conditioned on the test input $\xtest_n$ and $z$, which is compared with the test output $\ytest_n$ to obtain the test log-likelihood $\E_{z \sim \variq(z | x_{1:T}, y_{1:T})} [\log \truep(\ytest_n | \xtest_n, z)]$.
It would be ideal if we could analytically compute it, but if this is not the case, we may approximate it by the Monte Carlo estimation (sampling multiple $z$'s from $\variq(z | x_{1:T}, y_{1:T})$) or the maximum a posteriori estimation of $z$.

\subsection{Meta-Training}

During the meta-training phase, the model and the learner are meta-updated to maximize \eqref{eq:elbo:sup} or \ref{eq:elbo:unsup} with multiple CL episodes.
For each episode, the CL process in \S\ref{sec:method:cl} is used to obtain $\variq(z | \trainstream)$ with the learner.
In contrast to SGD-based MCL, our approach does not need to process the training stream sequentially.
If all the training examples are available, which is generally true during meta-training, we can feed them to the learner in parallel and combine the results with a batch inference rule instead of the sequential update rule.
With the Gaussian posterior, for example, we can use the following formula instead of \eqref{eq:post:seq} to produce the identical result:
\begin{align}
    \Lambda_T = \sum_{t=0}^T P_t,\hspace{3mm} \mu_T = \Lambda_T^{-1} \sum_{t=0}^T P_t \learnerz_t. \label{eq:post:batch}
\end{align}
Compared to SGD-based approaches requiring forward-backward passes for each example sequentially, the meta-training of our approach can benefit from parallel processors such as GPUs or TPUs.

Once the variational posterior $\variq(z | \trainstream)$ is obtained, we use Monte Carlo approximation for the expectation w.r.t. $\variq(z | \trainstream)$ \citep{VAE}.
For the Gaussian posterior, we can utilize the reparameterization trick \citep{VAE} to sample $z$ that allows backpropagation:
\begin{align}
    z = \mu_T + \Lambda_T^{-1/2} \epsilon,\hspace{3mm} \epsilon \sim \gN(0, I).
\end{align}
Conditioned on $z$, we run the model on the training and test examples to compute the first term in \eqref{eq:elbo:sup} or \ref{eq:elbo:unsup}.
This term encourages the cooperation between the model and the learner to increase the likelihood of the data.
The second term is the Kullback-Leibler (KL) divergence between the variational posterior $\variq(z | \trainstream)$ and the prior $\truep(z)$, which can be regarded as a regularization term.
We set the prior to be the same exponential family distribution, e.g., the unit Gaussian for the Gaussian posterior, which enables an analytical computation of the KL divergence.
Finally, the last term $\log \truep(\trainstream)$ is a constant that can be ignored for optimization purposes.

After \eqref{eq:elbo:sup} or \ref{eq:elbo:unsup} is computed for an episode or a batch of episodes, we perform a meta-update on the model and the learner with an SGD algorithm, backpropagating through the entire episode.
Unlike existing SGD-based MCL methods \citep{OML, ANML}, we do not need to calculate any second-order gradients, which is a significant advantage for scalability.

\begin{table*}[t!]
    \caption{Summary of the special cases of SB-MCL.}
    \label{tab:special}
    \vskip 3pt
    \centering
    \small
    \begin{tabular}{lllll}
    \toprule
    Method & Domain & Model structure & $z$ & $\variq(z | \trainstream)$ \\
    \midrule
    GeMCL & Classification & Encoder + GMM & GMM param. & Per-class Gaussian \\
    PN & Classification & Encoder + GMM & GMM param. & Per-class isotropic Gaussian \\
    ALPaCA & Regression & Encoder + Linear model & Linear model param. & Matrix normal \\
    \midrule
    SB-MCL & Any domain & Any model & Any auxiliary input & An exponential family distribution \\
    \bottomrule
    \end{tabular}
\end{table*}

\subsection{Existing Special Cases of SB-MCL}
\label{sec:method:special}

Several prior works can be considered domain-specific special cases of SB-MCL.
We summarize the key characteristics in Table \ref{tab:special} and high-level descriptions in the following.

\textbf{GeMCL \citep{GeMCL}.}
GeMCL can be regarded as a specific instance of our framework in the image classification domain.
It utilizes a meta-learned neural network encoder to extract an embedding vector for each image.
During the training process, it maintains a Gaussian posterior for each class in the embedding space.
Each Gaussian posterior is updated by the sequential Bayesian update rule whenever an example for the corresponding class becomes available.
These Gaussians collectively form a Gaussian mixture model (GMM) within the embedding space.
At test time, each test image is converted into an embedding vector by the same encoder, and a class is predicted by inferring the mixture component of GMM.
To view GeMCL as an instance of SB-MCL, we consider the encoder as serving two roles: one as the learner and the other as a component of the model.
During training, the encoder is used as the learner to update the posterior $\variq(z | x_{1:t}, y_{1:t})$ where $z$ is the parameters of the GMM.
At test time, the encoder transforms the test inputs into embeddings as a model component, and the GMM classifies the embeddings with its parameters learned from the training phase.
\citet{GeMCL} also propose an MAP variant, which simply produces $\truep(\ytest_n | \xtest_n, \zmap)$ as the output.
This variant has simpler computation without significant performance drops.

\textbf{Prototypical Networks \citep{PN}.}
While GeMCL is a special case of SB-MCL, it can also be seen as a generalization of the Prototypical Network (PN), which was originally proposed as a meta-learning approach for few-shot classification.
Therefore, PN also falls under the SB-MCL family.
While GeMCL takes a fully Bayesian approach, PN simply averages the embeddings of each class to construct a prototype vector.
Since the average operation can be performed sequentially, PN can be readily applied to MCL settings.
We can simplify GeMCL to PN by assuming isotropic Gaussian posteriors and an uninformative prior \citep{GeMCL}.

\textbf{ALPaCA \citep{ALPaCA}.}
Originally proposed as a meta-learning approach for online regression problems, ALPaCA attaches a linear model on top of a meta-learned neural network encoder, symmetrical to PN or GeMCL that attaches a GMM for classification.
In ALPaCA, the latent variable $z$ is the weight matrix of the linear model, whose posterior is assumed to have the matrix normal distribution.
Due to the similar streaming settings of online and continual learning, we can apply ALPaCA to MCL regression settings with minimal modifications.

\subsection{Converting Arbitrary Models for SB-MCL}
\label{sec:method:generic}

All the prior works in the previous section share a similar architecture: a meta-learned encoder followed by a simple statistical model.
This configuration can be ideal if the output type is suitable for the statistical model, allowing analytic computation of the posterior.
However, it is hard to apply such architectures to domains with more complex output formats or unsupervised settings where the output variable does not exist.

On the other hand, we can apply SB-MCL to almost any existing model architectures or domains, since the only modification is to be conditioned on some $z$ whose posterior is modeled with the exponential family.
Once the model is modified, a learner is added to digest the training stream into the variational posterior of $z$.
It may share most of its parameters with the model.

While there are infinitely many ways to implement such modifications, we currently focus on perhaps the simplest approach and leave exploring more sophisticated architectures for future work.
In our experiments, we define $z$ to be a 512-dimensional factorized Gaussian variable, which is injected into the model as an auxiliary input.
If the model structure follows an encoder-decoder architecture, we concatenate $z$ with the encoder output and pass the result to the decoder.
It should be noted that, despite its simplicity, a high-dimensional Gaussian can be surprisingly versatile when properly combined with neural networks.
This has also been demonstrated by generative models, such as VAEs \citep{VAE} or GANs \citep{GAN}, where neural networks transform a unit Gaussian variable into realistic images.
While their choice of Gaussian is motivated by the convenience of sampling, ours is motivated by its robustness to forgetting.

\section{Related Work}
\label{sec:related_work}

\textbf{SGD-Based MCL.}
OML \citep{OML} employs a small multi-layer perceptron (MLP) with MAML \citep{MAML} on top of a meta-learned encoder.
In the inner loop of OML, the encoder remains fixed while the MLP is updated by sequentially learning each training example via SGD.
After training the MLP in the inner loop, the entire model is evaluated on the test set to produce the meta-loss.
Then, the gradient of the meta-loss is computed with respect to the encoder parameters and the initial parameters of the MLP to update them.
Inspired by OML, ANML \citep{ANML} is another MCL method for image classification that introduces a component called neuromodulatory network.
Its sigmoid output is multiplied to the encoder output to adaptively gate some features depending on the input.
For a detailed survey of MCL and other combinations of meta-learning and CL, we refer the reader to \citet{MCL-Survey}.

\textbf{Continual Learning as Sequence Modeling (CL-Seq).}
More recently, \citet{CL-Seq} pointed out that CL is inherently a sequence modeling problem; predicting the target $\tilde y$ of a test input $\tilde x$ after a training stream $((x_1, y_1), ..., (x_T, y_T))$ is equivalent to predicting the next token $\tilde y$ that comes after prompt $(x_1, y_1, ..., x_T, y_T, \tilde x)$.
From this perspective, forwarding the training stream through an autoregressive sequence model and updating its internal state, which has been called \emph{in-context learning} in the language modeling literature \citep{GPT3}, can be considered CL.
Within MCL settings, the sequence model can be meta-trained on multiple CL episodes to perform CL.
They demonstrate that Transformer \citep{Transformer} and their efficient variants \citep{LinearTF, Performer} achieve significantly better scores compared to SGD-based approaches.

\textbf{Neural Processes.}
While motivated by different objectives, intriguing similarities can be identified between the supervised version of SB-MCL (\eqref{eq:elbo:sup}) and the neural process (NP) literature \citep{CNP,NP}.
NP was initially proposed to solve the limitations of Gaussian processes, such as the computational cost and the difficulties in the prior design.
It can also be considered a meta-learning approach that learns a functional prior and has been applied as a solution to the meta-learning domain \citep{ML-PIP}.
Since NPs are rooted in stochastic processes, one of their primary design considerations is exchangeability: the model should produce the same result regardless of the order of the training data.
To achieve exchangeability, NPs independently encode each example and aggregate them into a single variable with a permutation-invariant operation, such as averaging, and pass it to the decoder.
While our sequential Bayesian update of an exponential family posterior is initially inspired by the Fisher-Darmois-Koopman-Pitman theorem, it also ensures exchangeability.
\citet{Volpp21BayesContext} propose an aggregation scheme for NPs based on Bayesian principles and even suggest the possibility of sequential update, but they do not connect it to CL.
To the best of our knowledge, the only connection between NPs and MCL is CNAP \citep{CNAP}, but it is a domain-specific architecture designed for image classification.

\section{Experiments}

We demonstrate the efficacy of our framework on a wide range of domains, including both supervised and unsupervised tasks.
We also provide PyTorch \citep{PyTorch} code, ensuring the reproducibility of all experiments.
Due to page limitations, we present only the most essential information; for further details, please refer to the code.

\subsection{Methods}

\textbf{SGD-Based MCL.}
Due to its simplicity and generality, we test OML \citep{OML} as a representative baseline of SGD-based MCL.
Although it was originally proposed for classification and simple regression, \citet{CL-Seq} introduce an encoder-decoder variant of OML by stacking a MAML MLP block between the encoder and decoder, which can be used for other domains.
As the main computational bottleneck of OML is the second-order gradient computation, we also test its first-order approximation (OML-Rep), following Reptile \citep{Reptile}.

\textbf{CL-Seq.}
We test Transformer (TF; \citealp{Transformer}) and Linear Transformer (Linear TF; \citealp{LinearTF}) imported from the implementation of \citet{CL-Seq}.
In the case of TF, the computational cost keeps increasing as it learns more examples, which has been criticized as a major drawback limiting its scalability \citep{Tay2020EfficientTA}.
On the other hand, Linear TF maintains a constant computational cost like other baselines and our SB-MCL, but its performance falls behind TF \citep{CL-Seq}.

\textbf{Offline and Online Learning.}
Although our work focuses on MCL, a significant number of non-meta-CL methods have been proposed.
To provide a reference point to them, we report the offline and online learning scores, which are generally considered the upper bound of CL and online CL performance \citep{SI, Farajtabar20Orthogonal}.
For offline learning, we train a model from scratch for an unlimited number of SGD steps with mini-batches uniformly sampled from the entire training stream.
Since the model is usually overfitted to the training set, we report the best test score achieved during training.
For online learning, we randomly shuffle the training stream to be a stationary stream, train a model from scratch for one epoch, and measure the final test score.
Note that MCL methods can outperform offline and online learning since they can utilize a large meta-training set, unlike CL methods \citep{CL-Seq}.

\textbf{The SB-MCL Family (Ours).}
We test the special cases of SB-MCL in Table \ref{tab:special} for their respective domains, i.e., GeMCL for image classification, ALPaCA for simple regression, and the generic variant with the factorized Gaussian variable (\S\ref{sec:method:generic}) for others.
GeMCL and ALPaCA support the analytic calculation of posterior predictive distribution during testing.
For the generic cases, we impose 512D factorized Gaussian on $\variq(z | \trainstream)$ and sample $z$ five times to approximate $\E_{z \sim \variq(z | \trainstream)} [\truep(\ytest_n | \xtest_n, z)]$.
In Appendix \ref{app:values}, we also report the scores of its MAP variant that simply produces $\truep(\ytest_n | \xtest_n, \zmap)$.
The scores of MAP estimation are nearly the same as those of Monte Carlo estimation.

\subsection{Benchmarks}
\label{sec:exp:benchmark}

Our experimental settings are mainly based on those of \citet{CL-Seq}.
As the popular Omniglot dataset \citep{Omniglot} causes severe meta-overfitting due to its small size (1.6K classes / 32K images), they repurpose CASIA \citep{CASIA} and MS-Celeb-1M \citep{MSCeleb} datasets for MCL.
CASIA is a Chinese handwriting dataset that comprises 3.9M images of 7.4K character types, while MS-Celeb-1M contains 10M images of 100K celebrities.
Using these datasets, \citet{CL-Seq} test various types of supervised learning benchmarks, including both classification and regression.
Each class (e.g., character type or celebrity identity) is defined as a distinct task.
High-level descriptions of each benchmark are provided below.
We also provide visual illustrations of the model architectures used for each benchmark in Appendix \ref{app:arch}.

\textbf{Image Classification.}
We conduct experiments with the Omniglot, CASIA, and Celeb datasets, following the setups of \citet{CL-Seq}.
All the methods share the same CNN encoder with five convolutional layers.
GeMCL is compared as an instance of SB-MCL.

\textbf{Sine Regression.}
We adopt the synthetic sine wave regression setting from \citet{CL-Seq}.
ALPaCA is tested as an instance of SB-MCL.

\textbf{Image Completion (Compl.).}
$x$ and $y$ are an image's top and bottom halves, and each class is defined as a task.
We use the convolutional encoder-decoder architecture from \citet{CL-Seq}.
In the case of SB-MCL, we use the factorized Gaussian posterior and introduce another five-layer convolutional encoder for the learner, which produces $\variq(x, y | z)$ from a full training image.
The model's decoder is slightly modified to take the concatenation of the encoder's output and $z$ as input.

\textbf{Rotation Prediction.}
A model is given a randomly rotated image $x$ and tasked to predict the rotation angle $y$.
Although the rotation angle is not high-dimensional, we use the generic supervised SB-MCL architecture as in the image completion task.
This is due to the objective function, which is defined as $1 - \cos(y-\hat y)$ and cannot be used for analytically computing the posterior of the linear model in ALPaCA.
For the architecture, we use a convolutional encoder followed by an MLP output module.
For the learner in SB-MCL, we share the same encoder in the model for encoding $x$ and introduce a new MLP to encode $y$.
These two encoders' outputs are concatenated and fed to another MLP to produce $\variq(x, y | z)$.

\begin{table}[t]
    \caption{
        Classification results in the error rate ($\downarrow$).
    }
    \centering
    \small
    \label{tab:clf}
    \vskip 3pt
    \begin{tabular}{ccccccccc}
        \toprule
        Method & Omniglot & CASIA & Celeb \\
        \midrule
        Offline & $.300^{\pm .055}$ & $.345^{\pm .045}$ & $.625^{\pm .065}$ \\
        Online & $.800^{\pm .055}$ & $.963^{\pm .020}$ & $.863^{\pm .037}$ \\
        \midrule
        OML & $.046^{\pm .002}$ & $.015^{\pm .001}$ & $.331^{\pm .006}$ \\
        OML-Rep & $.136^{\pm .005}$ & $.057^{\pm .003}$ & $.660^{\pm .012}$ \\
        TF & $.014^{\pm .001}$ & $.004^{\pm .000}$ & $\mathbf{.228}^{\pm .003}$ \\
        Linear TF & $.125^{\pm .016}$ & $.006^{\pm .000}$ & $.229^{\pm .003}$ \\
        SB-MCL & $\mathbf{.008}^{\pm .000}$ & $\mathbf{.002}^{\pm .000}$ & $.231^{\pm .004}$ \\
        \bottomrule
    \end{tabular}
\end{table}
\begin{table}[t]
    \caption{
        Regression results in the loss ($\downarrow$).
    }
    \label{tab:reg}
    \vskip 3pt
    \centering
    \small
    \begin{tabular}{c@{\hspace{1.8mm}}c@{\hspace{1.8mm}}c@{\hspace{1.8mm}}c@{\hspace{1.8mm}}c@{\hspace{1.8mm}}c@{\hspace{1.8mm}}c@{\hspace{1.8mm}}c@{\hspace{1.8mm}}c}
        \toprule
        \multirow{2}{*}{Method} & \multirow{2}{*}{Sine} & CASIA & CASIA & Celeb \\
        &  & Compl. & Rotation & Compl. \\
        \midrule
        Offline & $.0045^{\pm .0003}$ & $.146^{\pm .009}$ & $.544^{\pm .045}$ & $.160^{\pm .008}$ \\
        Online & $.5497^{\pm .0375}$ & $.290^{\pm .023}$ & $1.079^{\pm .081}$ & $.284^{\pm .017}$ \\
        \midrule
        OML & $.0164^{\pm .0007}$ & $.105^{\pm .000}$ & $.052^{\pm .002}$ & $.099^{\pm .000}$ \\
        OML-Rep & $.0271^{\pm .0012}$ & $.104^{\pm .000}$ & $.050^{\pm .002}$ & $.105^{\pm .000}$ \\
        TF & $\mathbf{.0009}^{\pm .0001}$ & $\mathbf{.097}^{\pm .000}$ & $\mathbf{.034}^{\pm .001}$ & $\mathbf{.094}^{\pm .000}$ \\
        Linear TF & $.0031^{\pm .0002}$ & $.101^{\pm .000}$ & $.068^{\pm .002}$ & $.097^{\pm .000}$ \\
        SB-MCL & $.0011^{\pm .0002}$ & $.100^{\pm .001}$ & $.039^{\pm .001}$ & $.096^{\pm .000}$ \\
        \bottomrule
    \end{tabular}
\end{table}
\begin{table}[t]
    \caption{
        Results of deep generative models in the loss ($\downarrow$).
    }
    \label{tab:gen}
    \vskip 3pt
    \centering
    \small
    \begin{tabular}{ccccccccc}
      \toprule
      \multirow{2}{*}{Method} & CASIA & CASIA & Celeb \\
 & VAE & DDPM & DDPM \\
\midrule
Offline & $.664^{\pm .018}$ & $.0451^{\pm .0022}$ & $.0438^{\pm .0019}$ \\
Online & $.862^{\pm .009}$ & $.1408^{\pm .0032}$ & $.2124^{\pm .0025}$ \\
\midrule
OML & $.442^{\pm .003}$ & $.0353^{\pm .0001}$ & $.0308^{\pm .0003}$ \\
OML-Rep & $.454^{\pm .000}$ & $.0353^{\pm .0001}$ & $.0307^{\pm .0004}$ \\
SB-MCL & $\mathbf{.428}^{\pm .001}$ & $\mathbf{.0345}^{\pm .0001}$ & $\mathbf{.0302}^{\pm .0004}$ \\
      \bottomrule
    \end{tabular}
\end{table}

\textbf{Deep Generative Modeling.}
As the first in MCL research, we evaluate MCL performance with deep generative models.
We evaluate unsupervised learning performances with two types of deep generative models: variational autoencoder (VAE; \citealp{VAE}) and denoising diffusion probabilistic models (DDPM; \citealp{DDPM}).
We use a simple convolutional encoder-decoder architecture for VAE and a U-Net encoder-decoder architecture for DDPM following \citet{DDPM}.
In SB-MCL, we use a separate encoder for the learner, and $z$ is injected into the model by concatenating it with the decoder's input.
For OML, we replace the encoder's last MLP and the decoder's first MLP with a MAML MLP.
Transformers are not tested in this setting since they are not straightforward to be combined with deep generative models.

\textbf{Evaluation Scheme.}
In all MCL experiments, we meta-train the methods in a 10-task 10-shot setting: each training stream is a concatenation of 10 tasks with 10 examples each.
We primarily evaluate their performance in a meta-test set with the same task-shot setting, while also measuring the generalization capability on other meta-testing setups.
The hyperparameters are tuned to maximize the performance in the 10-task 10-shot settings.
We report classification errors for the classification benchmarks and losses for others.
Therefore, lower scores are always better.
For each experiment, we report the average and the standard deviation of five runs.
Within each MCL run, we calculate the average score from 512 CL episodes sampled from the meta-test set.
For offline and online learning, which do not involve any meta-training, we sample an episode from the meta-test set, train the model on the training set, and measure the test score.
We repeat this process 20 times and report the average and standard error of the mean.

\subsection{Results and Analyses}
\label{sec:exp:result}

\begin{figure}[h]
    \centering
    \begin{subfigure}[b]{\linewidth}
        \includegraphics[width=\linewidth]{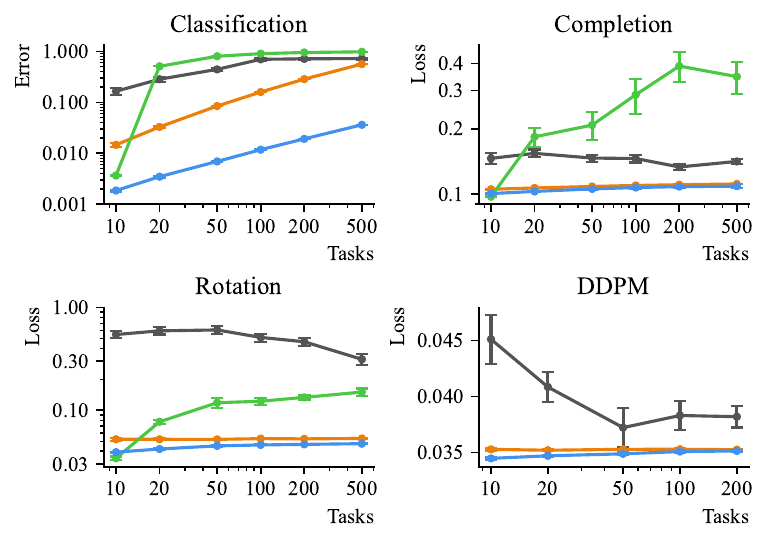}
        \caption{More tasks}
        \label{fig:longer:task}
    \end{subfigure}
    \begin{subfigure}[b]{\linewidth}
        \includegraphics[width=\linewidth]{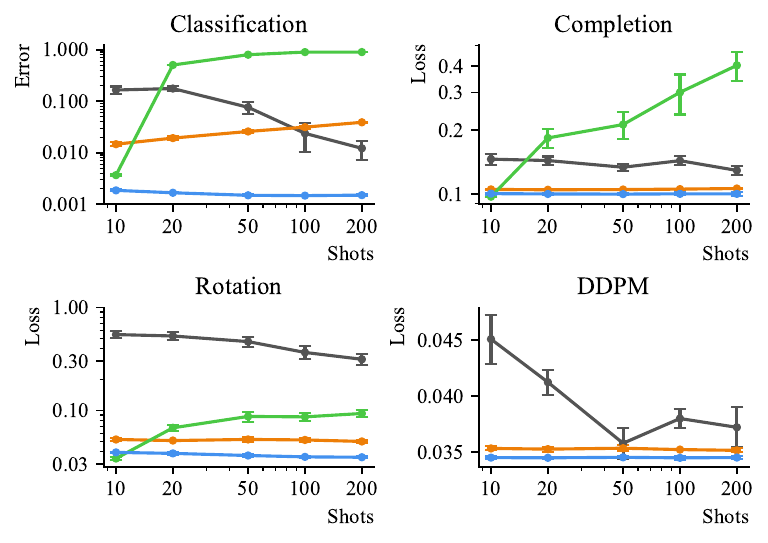}
        \caption{More shots}
        \label{fig:longer:shot}
    \end{subfigure}
    \includegraphics[width=\linewidth, trim={70pt 495pt 70pt 135pt}, clip]{figures/main.pdf}
    \caption{
        Generalization to longer training streams with more tasks and shots after meta-training with 10-task 10-shot on CASIA.
    }
    \label{fig:longer}
\end{figure}

We present our classification, regression, and deep generative modeling results in Table \ref{tab:clf}, \ref{tab:reg}, and \ref{tab:gen}, respectively.
\figref{fig:longer} compares the generalization abilities in longer training streams, while Table \ref{tab:ood} summarizes generalization to a different dataset.
For qualitative examples and more extensive results, please refer to Appendix \ref{app:qualitative} and \ref{app:values}.
We discuss several notable characteristics of our SB-MCL that can be observed in the experiments.

\textbf{Strong CL Performance.}
In the classification, regression, and generation experiments (Table \ref{tab:clf}-\ref{tab:gen}), the SB-MCL family significantly outperforms SGD-based approaches and Linear TF.
Its performance is comparable to TF, whose per-example computational cost constantly grows with the number of learned examples.

\textbf{Stronger Generalization Ability.}
When meta-tested on longer training streams (\figref{fig:longer}) or a different dataset (Table \ref{tab:ood}), SB-MCL achieves substantially better scores than all the other baselines.
Especially, TF's performance degrades catastrophically due to its poor length generalization ability, which is a well-known limitation of TF \citep{TFLength}.
Another interesting point is that TF and OML's performance can degrade even when provided with more shots and the same number of tasks as presented in \figref{fig:longer:shot}.
This may seem counterintuitive, as providing more information about a task without adding more tasks should generally be beneficial.
In SGD-based MCL, however, the longer training stream results in more SGD updates, which can exacerbate forgetting.
TF's performance deteriorates even more dramatically due to length generalization failure.
On the other hand, the SB-MCL family demonstrates a remarkable level of robustness in many-shot settings.
As the number of shots increases, their performance even improves slightly. 
This observation aligns with our formulation.
Since our posterior follows an exponential family distribution with fixed-sized sufficient statistics, maintaining the same number of tasks while increasing the number of shots serves only to enhance the accuracy of the variational posterior.

\begin{table}[t]
    \caption{
        Generalization to another dataset. Meta-test scores on Omniglot after meta-training on CASIA.
    }
    \label{tab:ood}
    \vskip 3pt
    \centering
    \small
    \begin{tabular}{c@{\hspace{2mm}}c@{\hspace{2mm}}c@{\hspace{2mm}}c@{\hspace{2mm}}c@{\hspace{2mm}}c@{\hspace{2mm}}c@{\hspace{2mm}}c@{\hspace{2mm}}c}
      \toprule
      Method & Classification & Rotation & VAE & DDPM \\
\midrule
OML & $.445^{\pm .020}$ & $.856^{\pm .074}$ & $.227^{\pm .002}$ & $.027^{\pm .000}$ \\
OML-Rep & $.496^{\pm .023}$ & $.736^{\pm .010}$ & $.244^{\pm .001}$ & $.027^{\pm .000}$ \\
TF & $.088^{\pm .010}$ & $.850^{\pm .015}$ & -- & -- \\
Linear TF & $.102^{\pm .011}$ & $.931^{\pm .031}$ & -- & -- \\
SB-MCL & $\mathbf{.023}^{\pm .001}$ & $\mathbf{.640}^{\pm .012}$ & $\mathbf{.219}^{\pm .001}$ & $\mathbf{.026}^{\pm .000}$ \\
      \bottomrule
    \end{tabular}
\end{table}

\begin{table}[t]
    \caption{
        Meta-training time comparison.
        We report the time required to meta-train for 50K steps with a single A40 GPU.
    }
    \label{tab:time}
    \vskip 3pt
    \centering
    \small
    \begin{tabular}{ccccccccc}
        \toprule
        Method & OML & TF & SB-MCL \\
        \midrule
        Classification & 6.5 hr & 1.2 hr & \textbf{40 min} \\
        Completion & 16.5 hr & 1.4 hr & \textbf{1.2 hr} \\
        VAE & 19 hr & N/A & \textbf{1.2 hr} \\
        DDPM & 5 days & N/A & \textbf{8 hr} \\
        \bottomrule
    \end{tabular}
\end{table}

\textbf{Superior Efficiency.}
In Table \ref{tab:time}, we compare the meta-training time of the SB-MCL family against OML and TF.
First of all, SB-MCL and TF are significantly faster than OML, which does not support parallel training.
Parallel training is essential for utilizing parallel processors like GPUs for efficient meta-training.
SB-MCL is faster than TF in all the benchmarks, demonstrating its superior efficiency due to the constant computational cost of the Bayesian update.

\textbf{CL as a Matter of Representational Capacity.}
By design, SB-MCL yields the same results regardless of whether the training data is provided sequentially or not; in other words, \emph{no forgetting} is theoretically guaranteed.
This unique property enables new approaches to CL; instead of dealing with the complex learning dynamics of SGD on a non-stationary training stream, we can focus on maximizing the representational capacity.
This includes designing better/bigger architectures and collecting more data, just like solving ordinary deep-learning problems in offline settings.
Note that this has not been possible with SGD-based approaches since their CL performance is not necessarily aligned with the representational capacity due to the complicated dynamics of forgetting.

\section{Conclusion}
This work introduces a general MCL framework that combines the exponential family's robustness to forgetting and the flexibility of neural networks.
Its superior performance and efficiency are empirically demonstrated in diverse domains.
Unifying several prior works under the same framework, we aim to establish a solid foundation for the future sequential Bayesian approaches in the field of MCL.
As discussed in \S\ref{sec:exp:result}, our framework reframes CL's forgetting issue as a matter of representational capacity.
This allows us to focus on the architectural aspect, rather than the optimization aspect of preventing forgetting.
Designing neural architectures for interacting with the exponential family posterior can be an exciting avenue for further research.
Collecting new datasets for MCL also arises as an important future direction.
While our method can benefit from large-scale data, few datasets are available for MCL research at the moment.
We believe our approach can enable interesting applications when combined with appropriate datasets.

\section*{Limitation}

While our framework demonstrates strong performance across various MCL tasks, it faces a fundamental limitation due to the assumption of an exponential family posterior.
The equivalence between the sequential update rule and batch learning, while preventing forgetting, completely disregards the order of training data.
This is acceptable and even beneficial when data order is irrelevant, as observed in the standard CL benchmarks used in our experiments.
However, in real-world applications, the sequence of training data can be crucial. For instance, training data may be organized into a curriculum where acquiring new knowledge depends on previously learned information.
In such scenarios, our framework may not be the optimal choice.

Our research began with the constraint of maintaining a constant memory size throughout the learning process.
The Fisher-Darmois-Koopman-Pitman theorem indicates that only an exponential family posterior can prevent forgetting under this constraint.
By relaxing this constraint, we could explore more flexible, non-parametric posterior distributions.
We propose this as an intriguing direction for future research.

\section*{Impact Statement}

This paper contributes to the field of machine learning, specifically in continual learning.
While recognizing the potential societal consequences of our work, we conclude that no particular aspects demand specific highlighting.

\section*{Acknowledgements}

This work was supported by Samsung Advanced Institute of Technology and the Institute of Information \& communications Technology Planning \& Evaluation (IITP) grants funded by the Korea government (MSIT) (No.~RS-2022-II220156, Fundamental research on continual meta-learning for quality enhancement of casual videos and their 3D metaverse transformation; No.~RS-2019-II191082, SW StarLab; No.~RS-2021-II211343, Artificial Intelligence Graduate School Program (Seoul National University)).

\bibliography{icml2024_seq_bayes_mcl}
\bibliographystyle{icml2024}

\newpage
\appendix
\onecolumn
\section{Variational Bound Derivation}
\label{app:derivation}

The derivation of the variational bound for supervised learning setup (\eqref{eq:elbo:sup}) is as follows:
\begin{align}
&\log \truep(\ytest_{1:N} | \xtest_{1:N}, \trainstream) \nonumber \\
&= -\log \truep(z | \ytest_{1:N}, \xtest_{1:N}, \trainstream) + \log \truep(\ytest_{1:N}, z | \xtest_{1:N}, \trainstream) \nonumber \\
&= \E_{z \sim \variq(z | \trainstream)} \left[
    \log \variq(z | \trainstream) - \log \truep(z | \ytest_{1:N}, \xtest_{1:N}, \trainstream) + \log \truep(\ytest_{1:N}, z | \xtest_{1:N}, \trainstream) - \log \variq(z | \trainstream)
    \right] \nonumber \\
&= \kldiv \left( \variq(z | \trainstream) \, \middle\| \, \truep(z | \ytest_{1:N}, \xtest_{1:N}, \trainstream) \right) 
    + \E_{z \sim \variq(z | \trainstream)} \left[ \log \truep(\ytest_{1:N}, z | \xtest_{1:N}, \trainstream) - \log \variq(z | \trainstream) \right] \nonumber \\
&\geq \E_{z \sim \variq(z | \trainstream)} \left[ \log \truep(\ytest_{1:N}, z | \xtest_{1:N}, \trainstream) - \log \variq(z | \trainstream) \right] \nonumber \\
&= \E_{z \sim \variq(z | \trainstream)} \left[
    \log \truep(\ytest_{1:N} | z, \xtest_{1:N})
    + \log \truep(z | \xtest_{1:N}, \trainstream)
    - \log \variq(z | \trainstream)
    \right] \label{eq:elbo:interim} \\
&= \E_{z \sim \variq(z | \trainstream)} \left[
    \log \truep(\ytest_{1:N} | z, \xtest_{1:N})
    + \log \truep(\trainstream | z, \xtest_{1:N}) + \log \truep(z | \xtest_{1:N}) - \log \truep(\trainstream | \xtest_{1:N}) \right. \nonumber \\
    &\left. \hspace{62pt} - \log \variq(z | \trainstream)
    \right] \nonumber \\
&= \E_{z \sim \variq(z | \trainstream)} \left[
    \log \truep(\ytest_{1:N} | z, \xtest_{1:N})
    + \log \truep(\trainstream | z) + \log \truep(z) - \log \truep(\trainstream)
    - \log \variq(z | \trainstream)
    \right] \nonumber \\
&= \E_{z \sim \variq(z | \trainstream)} \left[
    \log \truep(\ytest_{1:N} | z, \xtest_{1:N})
    + \log \truep(\trainstream | z)
    \right]
    - \kldiv \left( \variq(z | \trainstream) \, \middle\| \, \truep(z) \right)
    - \log \truep(\trainstream) \nonumber \\
&= \E_{z \sim \variq(z | \trainstream)} \left[
    \sum_{n=1}^N \log \truep(\ytest_n | \xtest_n, z) + \sum_{t=1}^T \log \truep(y_t | x_t, z)
    \right] \nonumber \\
    &\, \hspace{12pt} - \kldiv \left( \variq(z | \trainstream) \, \middle\| \, \truep(z) \right) - \underbrace{\log \truep(\trainstream)}_{\mathrm{const.}} \nonumber
\end{align}

We can derive a similar bound for unsupervised settings (\eqref{eq:elbo:unsup}):
\begin{align}
    &\log \truep(\xtest_{1:N} | \trainstream) \nonumber \\
    &= -\log \truep(z | \xtest_{1:N}, \trainstream) + \log \truep(\xtest_{1:N}, z | \trainstream) \nonumber \\
    &= \E_{z \sim \variq(z | \trainstream)} \left[
        \log \variq(z | \trainstream) - \log \truep(z | \xtest_{1:N}, \trainstream)
        + \log \truep(\xtest_{1:N}, z | \trainstream) - \log \variq(z | \trainstream)
        \right] \nonumber \\
    &= \kldiv \left(\variq(z | \trainstream) \, \middle\| \, \truep(z | \trainstream) \right)
        + \E_{z \sim \variq(z | \trainstream)} \left[
            \log \truep(\xtest_{1:N}, z | \trainstream) - \log \variq(z | \trainstream)
        \right] \nonumber \\
    &\geq \E_{z \sim \variq(z | \trainstream)} \left[
        \log \truep(\xtest_{1:N}, z | \trainstream) - \log \variq(z | \trainstream)
        \right] \nonumber \\
    &= \E_{z \sim \variq(z | \trainstream)} \left[
        \log \truep(\xtest_{1:N} | z, \trainstream) + \log \truep(z | \trainstream)
        - \log \variq(z | \trainstream)
        \right] \nonumber \\
    &= \E_{z \sim \variq(z | \trainstream)} \left[
        \log \truep(\xtest_{1:N} | z) + \log \truep(\trainstream | z) + \log \truep(z) - \log \truep(\trainstream)
        - \log \variq(z | \trainstream)
        \right] \nonumber \\
    &= \E_{z \sim \variq(z | \trainstream)} \left[
        \log \truep(\xtest_{1:N} | z) + \log \truep(\trainstream | z)
        \right]
        - \kldiv \left( \variq(z | \trainstream) \, \middle\| \, \truep(z) \right) - \log \truep(\trainstream) \nonumber \\
    &= \E_{z \sim \variq(z | \trainstream)} \left[
        \sum_{n=1}^N \log \truep(\xtest_n | z) + \sum_{t=1}^T \log \truep(x_t | z)
        \right]
        - \kldiv \left( \variq(z | \trainstream) \, \middle\| \, \truep(z) \right) - \underbrace{\log \truep(\trainstream)}_{\mathrm{const.}} \nonumber
\end{align}

It is noteworthy that Neural Process \citep{NP} instead approximates $\log \truep(z | \xtest_{1:N}, \trainstream)$ of Eq.~\ref{eq:elbo:interim} with $\log \variq(z | \xtest_{1:N}, \trainstream)$:
\begin{align}
&\E_{z \sim \variq(z | \trainstream)} \left[
    \log \truep(\ytest_{1:N} | z, \xtest_{1:N})
    + \log \truep(z | \xtest_{1:N}, \trainstream)
    - \log \variq(z | \trainstream)
    \right] \nonumber \\
&\approx \E_{z \sim \variq(z | \trainstream)} \left[
    \log \truep(\ytest_{1:N} | z, \xtest_{1:N})
    + \log \variq(z | \xtest_{1:N}, \trainstream)
    - \log \variq(z | \trainstream)
    \right] \nonumber \\
&= \E_{z \sim \variq(z | \trainstream)} \left[
    \sum_{n=1}^N \log \truep(\ytest_n | \xtest_n, z)
    \right]
    - \kldiv \left( \variq(z | \trainstream) \middle\| \variq(z | \xtest_{1:N}, \trainstream) \right) \nonumber
\end{align}
Since we can use the Bayes rule to convert $\log \truep(z | \xtest_{1:N}, \trainstream)$ into $\log \truep(\trainstream | z, \xtest_{1:N}) + \log \truep(z | \xtest_{1:N}) - \log \truep(\trainstream | \xtest_{1:N})$, which is subsequently reduced to $\log \truep(\trainstream | z) + \log \truep(z) - \log \truep(\trainstream)$ by conditional independence, such an approximation is not necessary.

\section{Architecture Diagrams}
\label{app:arch}

\begin{figure}[h!]
    \centering
    \includegraphics[page=3, width=\textwidth]{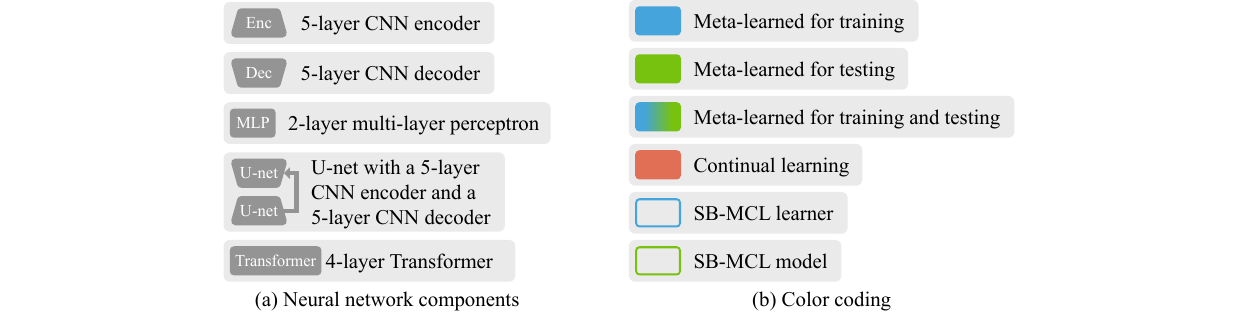}
    \caption{Notations for architecture diagrams.}
    \label{fig:arch:notation}
\end{figure}

\begin{figure}[h!]
    \centering
    \includegraphics[page=4, width=\textwidth]{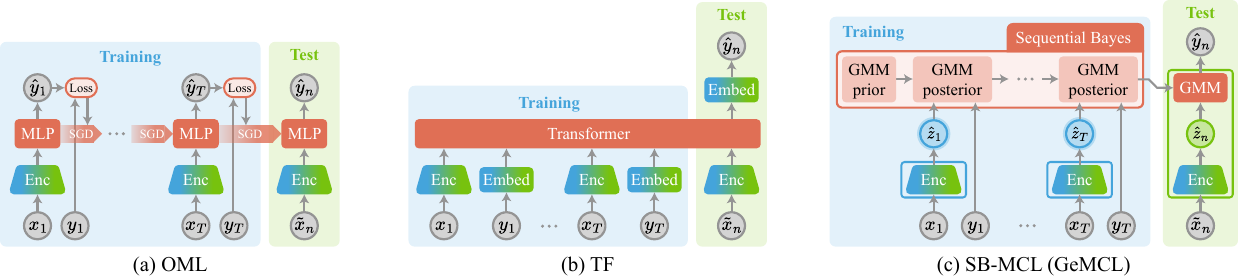}
    \caption{Architectures for classification experiments.}
    \label{fig:arch:clf}
\end{figure}

\begin{figure}[h!]
    \centering
    \includegraphics[page=5, width=\textwidth]{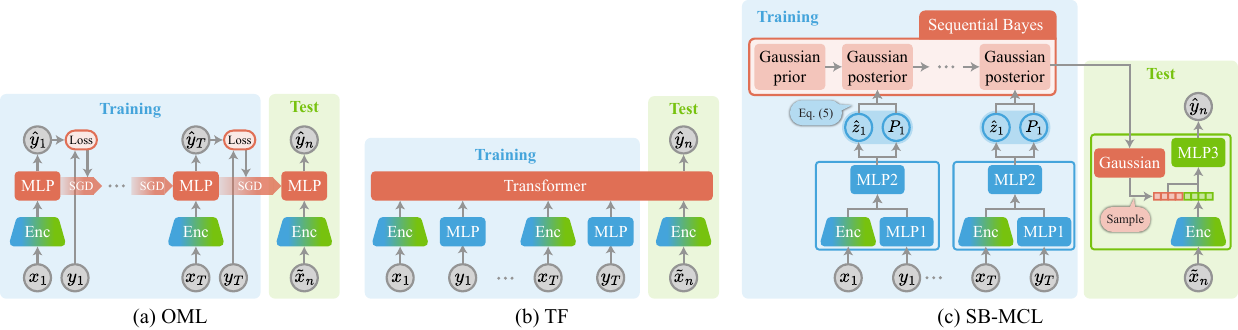}
    \caption{Architectures for rotation experiments.}
    \label{fig:arch:rot}
\end{figure}

\begin{figure}[h!]
    \centering
    \includegraphics[page=6, width=\textwidth]{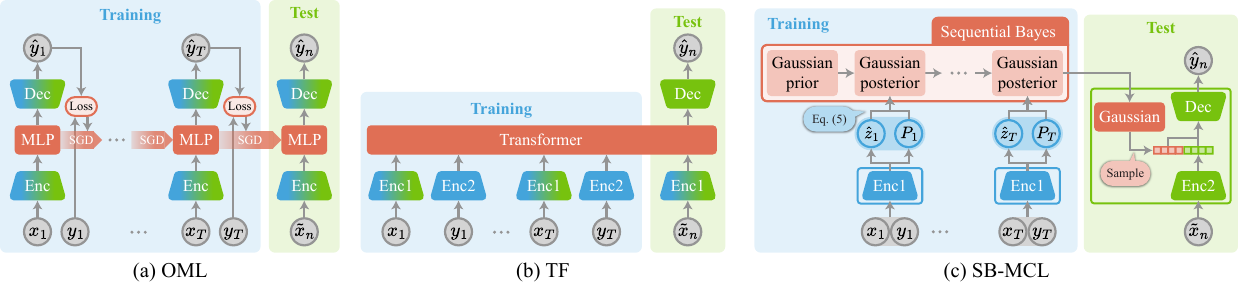}
    \caption{Architectures for completion experiments.}
    \label{fig:arch:comp}
\end{figure}

\begin{figure}[h!]
    \centering
    \includegraphics[page=7, width=\textwidth]{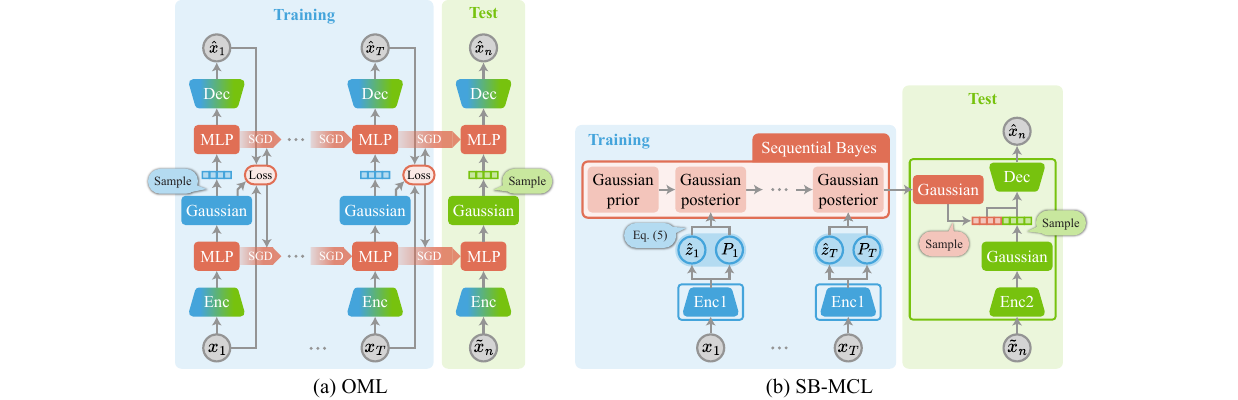}
    \caption{Architectures for VAE experiments.}
    \label{fig:arch:vae}
\end{figure}

\begin{figure}[h!]
    \centering
    \includegraphics[page=8, width=\textwidth]{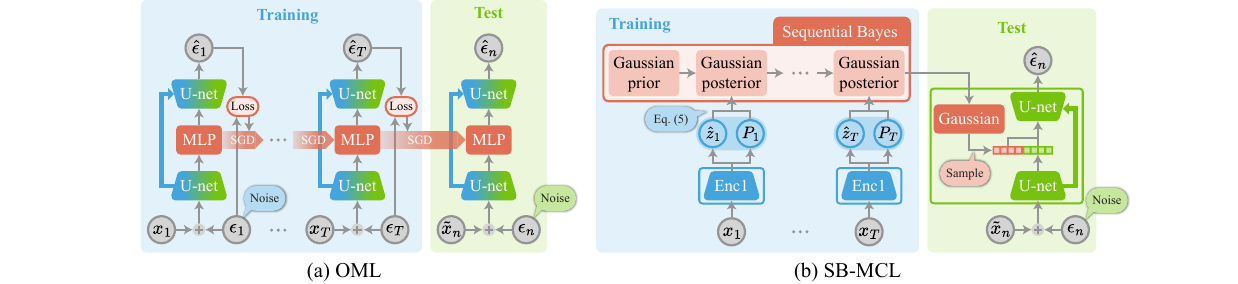}
    \caption{Architectures for DDPM experiments.}
    \label{fig:arch:ddpm}
\end{figure}

For better understanding of the architectures used in the experiments, we provide detailed diagrams for each experiment.
In \figref{fig:arch:notation}, we present the notations used in the architecture diagrams.
In \figref{fig:arch:clf}-\ref{fig:arch:ddpm}, we present the architectures used in the classification, rotation, completion, VAE, and DDPM experiments, respectively.

\section{Qualitative Examples of Deep Generative MCL}
\label{app:qualitative}

\begin{figure}[h]
    \centering
    \includegraphics[page=2, width=0.85\textwidth, trim={0 618pt 0 0}, clip]{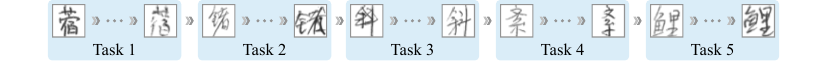}
    \caption{An example training stream from CASIA.}
    \label{fig:qual:train}
\end{figure}

\begin{figure}[h]
    \centering
    \includegraphics[page=2, width=0.85\textwidth, trim={0 338pt 0 280pt}, clip]{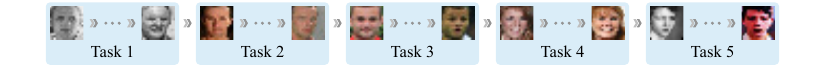}
    \caption{An example training stream from Celeb.}
    \label{fig:qual:train_celeb}
\end{figure}

\begin{figure}[h]
    \begin{minipage}[b]{0.49\textwidth}
        \centering
        \includegraphics[page=2, width=0.95\textwidth, trim={95pt 534pt 80pt 40pt}, clip]{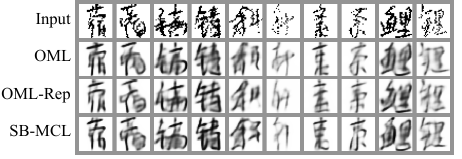}
        \caption{VAE reconstruction samples (CASIA).}
        \label{fig:qual:vae-recon}
    \end{minipage}
    \hfill
    \begin{minipage}[b]{0.49\textwidth}
        \begin{subfigure}[b]{\textwidth}
            \centering
            \includegraphics[page=2, width=182pt, trim={0 490pt 214pt 120pt}, clip]{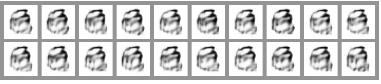}
            \caption{OML}
        \end{subfigure}
        \vspace{0mm}
        
        \begin{subfigure}[b]{\textwidth}
            \centering
            \includegraphics[page=2, width=182pt, trim={0 450pt 214pt 160pt}, clip]{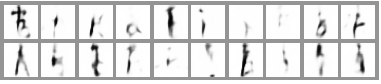}
            \caption{OML-Rep}
        \end{subfigure}
        \vspace{0mm}
        
        \begin{subfigure}[b]{\textwidth}
            \centering
            \includegraphics[page=2, width=182pt, trim={0 410pt 214pt 200pt}, clip]{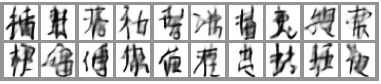}
            \caption{SB-MCL}
        \end{subfigure}
        \caption{VAE generation samples (CASIA).}
        \label{fig:qual:vae-gen}
    \end{minipage}
\end{figure}

\begin{figure}[h]
\begin{minipage}[b]{0.49\textwidth}
    \begin{subfigure}[b]{\textwidth}
        \centering
        \includegraphics[page=2, width=182pt, trim={214pt 490pt 0 120pt}, clip]{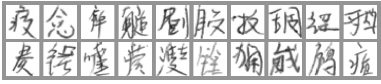}
        \caption{OML}
    \end{subfigure}
    \vspace{0mm}
    
    \begin{subfigure}[b]{\textwidth}
        \centering
        \includegraphics[page=2, width=182pt, trim={214pt 450pt 0 160pt}, clip]{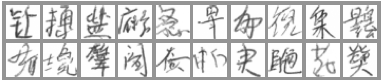}
        \caption{OML-Rep}
    \end{subfigure}
    \vspace{0mm}
    
    \begin{subfigure}[b]{\textwidth}
        \centering
        \includegraphics[page=2, width=182pt, trim={214pt 410pt 0 200pt}, clip]{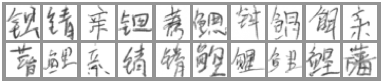}
        \caption{SB-MCL}
    \end{subfigure}
    \caption{DDPM generation samples (CASIA).}
    \label{fig:qual:casia-ddpm-gen}
\end{minipage}
\hfill
\begin{minipage}[b]{0.49\textwidth}
    \begin{subfigure}[b]{\textwidth}
        \centering
        \includegraphics[page=2, width=182pt, trim={214pt 290pt 0 320pt}, clip]{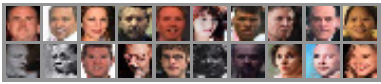}
        \caption{OML}
    \end{subfigure}
    \vspace{0mm}
    
    \begin{subfigure}[b]{\textwidth}
        \centering
        \includegraphics[page=2, width=182pt, trim={214pt 250pt 0 360pt}, clip]{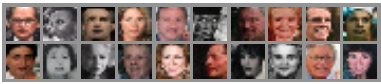}
        \caption{OML-Rep}
    \end{subfigure}
    \vspace{0mm}
    
    \begin{subfigure}[b]{\textwidth}
        \centering
        \includegraphics[page=2, width=182pt, trim={214pt 210pt 0 400pt}, clip]{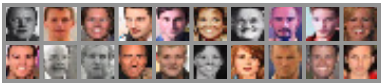}
        \caption{SB-MCL}
    \end{subfigure}
    \caption{DDPM generation samples (Celeb).}
    \label{fig:qual:celeb-ddpm-gen}
\end{minipage}

\end{figure}

In \figref{fig:qual:train}-\ref{fig:qual:celeb-ddpm-gen}, we present qualitative examples of the deep generative model experiments.
For VAEs, we use a binarized CASIA dataset for easier likelihood calculation, while using unmodified CASIA and MS-Celeb-1M datasets for DDPMs.
With each meta-trained MCL method, we train a VAE or DPMM on a 5-task 10-shot training stream in \figref{fig:qual:train} or \ref{fig:qual:train_celeb}, which are sampled from the meta-test set.
Then, we extract 20 generation samples for the VAE (\figref{fig:qual:vae-gen}) and the DDPM (\figref{fig:qual:casia-ddpm-gen} and \ref{fig:qual:celeb-ddpm-gen}).
For the VAE, we also visualize the reconstructions of the test images in \figref{fig:qual:vae-recon}.

Although the scores of OML and OML-Reptile are much worse than SB-MCL, the reconstruction results in \figref{fig:qual:vae-recon} do not seem to show a significant difference.
However, the generation results in \figref{fig:qual:vae-gen} of OML and OML-Reptile are not properly structured, showing that OML and OML-Reptile have difficulty in training VAE on a non-stationary stream.
On the other hand, the VAE with SB-MCL produces significantly better samples, demonstrating the effectiveness of our approach.

All the DDPM samples in \figref{fig:qual:casia-ddpm-gen} and \ref{fig:qual:celeb-ddpm-gen} are of much higher quality compared to VAE and are hard to distinguish from real images.
Since the DDPMs meta-learn general concepts from the large-scale meta-training set, they can produce high-fidelity images.
The key difference to notice is whether the DDPM has learned new knowledge from the training stream.
Since the training stream is from the meta-test set, it cannot produce the classes in the training stream unless it actually learns from it.
Among the samples from OML and OML-Reptile, it is hard to find the classes in the training stream, suggesting that they are producing samples from the meta-training distribution.
On the other hand, the DDPMs with SB-MCL produce samples remarkably similar to the ones in \figref{fig:qual:train} and \ref{fig:qual:train_celeb}.
This experiment confirms that SB-MCL can be an effective solution for modern deep generative models.

\section{Extended Experimental Results}
\label{app:values}

\begin{table}[h]
\centering
\caption{CASIA classification with more tasks.}
\vskip 0.1in
\small
\begin{tabular}{lrrrrrrr}
\toprule

\multirow{2}{*}{Method} & \multicolumn{6}{c}{Tasks} \\
 & 10 & 20 & 50 & 100 & 200 & 500 \\
\midrule
Offline & $.165^{\pm .028}$ & $.284^{\pm .033}$ & $.444^{\pm .038}$ & $.700^{\pm .038}$ & $.714^{\pm .034}$ & $.725^{\pm .031}$ \\
Online & $.963^{\pm .020}$ & $.925^{\pm .031}$ & $.963^{\pm .020}$ & $.963^{\pm .020}$ & $.963^{\pm .013}$ & $.970^{\pm .007}$ \\
\midrule
OML & $.015^{\pm .001}$ & $.033^{\pm .001}$ & $.085^{\pm .001}$ & $.159^{\pm .001}$ & $.286^{\pm .002}$ & $.564^{\pm .001}$ \\
OML-Rep & $.057^{\pm .003}$ & $.104^{\pm .002}$ & $.215^{\pm .004}$ & $.359^{\pm .002}$ & $.559^{\pm .005}$ & $.796^{\pm .003}$ \\
TF & $.004^{\pm .000}$ & $.510^{\pm .001}$ & $.804^{\pm .001}$ & $.903^{\pm .001}$ & $.952^{\pm .000}$ & $.980^{\pm .000}$ \\
SB-MCL & $.002^{\pm .000}$ & $.003^{\pm .000}$ & $.007^{\pm .000}$ & $.012^{\pm .000}$ & $.019^{\pm .000}$ & $.036^{\pm .000}$ \\
SB-MCL (MAP) & $.002^{\pm .000}$ & $.003^{\pm .000}$ & $.007^{\pm .000}$ & $.012^{\pm .000}$ & $.019^{\pm .000}$ & $.036^{\pm .000}$ \\
\bottomrule
\end{tabular}
\end{table}

\begin{table}[h]
\centering
\caption{CASIA classification with more shots.}
\vskip 0.1in
\small
\begin{tabular}{lrrrrrrr}
\toprule

\multirow{2}{*}{Method} & \multicolumn{5}{c}{Shots} \\
 & 10 & 20 & 50 & 100 & 200 \\
\midrule
Offline & $.165^{\pm .028}$ & $.176^{\pm .021}$ & $.076^{\pm .021}$ & $.024^{\pm .013}$ & $.012^{\pm .005}$ \\
Online & $.963^{\pm .020}$ & $.838^{\pm .032}$ & $.662^{\pm .041}$ & $.550^{\pm .074}$ & $.388^{\pm .065}$ \\
\midrule
OML & $.015^{\pm .001}$ & $.019^{\pm .001}$ & $.026^{\pm .002}$ & $.031^{\pm .002}$ & $.039^{\pm .001}$ \\
OML-Rep & $.057^{\pm .003}$ & $.066^{\pm .002}$ & $.083^{\pm .004}$ & $.101^{\pm .002}$ & $.121^{\pm .003}$ \\
TF & $.004^{\pm .000}$ & $.505^{\pm .001}$ & $.800^{\pm .000}$ & $.899^{\pm .001}$ & $.899^{\pm .000}$ \\
Linear TF & $.006^{\pm .000}$ & $.530^{\pm .010}$ & $.768^{\pm .028}$ & $.804^{\pm .031}$ & $.818^{\pm .038}$ \\
SB-MCL & $.002^{\pm .000}$ & $.002^{\pm .000}$ & $.001^{\pm .000}$ & $.001^{\pm .000}$ & $.002^{\pm .000}$ \\
SB-MCL (MAP) & $.002^{\pm .000}$ & $.002^{\pm .000}$ & $.001^{\pm .000}$ & $.001^{\pm .000}$ & $.001^{\pm .000}$ \\
\bottomrule
\end{tabular}
\end{table}

\begin{table}[h]
\centering
\caption{Sine classification with more tasks.}
\vskip 0.1in
\small
\begin{tabular}{lrrrrrrr}
\toprule

\multirow{2}{*}{Method} & \multicolumn{6}{c}{Tasks} \\
 & 10 & 20 & 50 & 100 & 200 & 500 \\
\midrule
Offline & $.005^{\pm .000}$ & $.004^{\pm .001}$ & $.005^{\pm .001}$ & $.008^{\pm .001}$ & $.036^{\pm .008}$ & $.198^{\pm .021}$ \\
Online & $.550^{\pm .037}$ & $.525^{\pm .032}$ & $.590^{\pm .030}$ & $.549^{\pm .031}$ & $.526^{\pm .022}$ & $.569^{\pm .013}$ \\
\midrule
OML & $.016^{\pm .001}$ & $.034^{\pm .002}$ & $.082^{\pm .001}$ & $.153^{\pm .002}$ & $.270^{\pm .000}$ & $.484^{\pm .002}$ \\
OML-Rep & $.027^{\pm .001}$ & $.054^{\pm .002}$ & $.115^{\pm .003}$ & $.201^{\pm .004}$ & $.335^{\pm .005}$ & $.559^{\pm .003}$ \\
TF & $.001^{\pm .000}$ & $.238^{\pm .020}$ & $.454^{\pm .011}$ & $.535^{\pm .011}$ & $.586^{\pm .013}$ & $.615^{\pm .006}$ \\
Linear TF & $.003^{\pm .000}$ & $.201^{\pm .011}$ & $.409^{\pm .011}$ & $.489^{\pm .006}$ & $.526^{\pm .003}$ & $.543^{\pm .002}$ \\
SB-MCL & $.001^{\pm .000}$ & $.002^{\pm .000}$ & $.007^{\pm .000}$ & $.020^{\pm .000}$ & $.065^{\pm .001}$ & $.228^{\pm .001}$ \\
\bottomrule
\end{tabular}
\end{table}

\begin{table}[h]
\centering
\caption{Sine classification with more shots.}
\vskip 0.1in
\small
\begin{tabular}{lrrrrrrr}
\toprule

\multirow{2}{*}{Method} & \multicolumn{5}{c}{Shots} \\
 & 10 & 20 & 50 & 100 & 200 \\
\midrule
Offline & $.005^{\pm .000}$ & $.003^{\pm .000}$ & $.003^{\pm .000}$ & $.002^{\pm .000}$ & $.002^{\pm .000}$ \\
Online & $.550^{\pm .037}$ & $.446^{\pm .031}$ & $.376^{\pm .031}$ & $.273^{\pm .018}$ & $.219^{\pm .017}$ \\
\midrule
OML & $.016^{\pm .001}$ & $.018^{\pm .001}$ & $.017^{\pm .001}$ & $.017^{\pm .001}$ & $.018^{\pm .001}$ \\
OML-Rep & $.027^{\pm .001}$ & $.027^{\pm .001}$ & $.027^{\pm .002}$ & $.027^{\pm .002}$ & $.027^{\pm .002}$ \\
TF & $.001^{\pm .000}$ & $.152^{\pm .030}$ & $.212^{\pm .044}$ & $.221^{\pm .034}$ & $.199^{\pm .039}$ \\
Linear TF & $.003^{\pm .000}$ & $.140^{\pm .012}$ & $.212^{\pm .017}$ & $.228^{\pm .026}$ & $.252^{\pm .022}$ \\
SB-MCL & $.001^{\pm .000}$ & $.001^{\pm .000}$ & $.001^{\pm .000}$ & $.001^{\pm .000}$ & $.001^{\pm .000}$ \\
\bottomrule
\end{tabular}
\end{table}

\begin{table}[h]
\centering
\caption{CASIA completion with more tasks.}
\vskip 0.1in
\small
\begin{tabular}{lrrrrrrr}
\toprule

\multirow{2}{*}{Method} & \multicolumn{6}{c}{Tasks} \\
 & 10 & 20 & 50 & 100 & 200 & 500 \\
\midrule
Offline & $.146^{\pm .009}$ & $.154^{\pm .006}$ & $.146^{\pm .005}$ & $.146^{\pm .006}$ & $.133^{\pm .005}$ & $.141^{\pm .004}$ \\
Online & $.290^{\pm .023}$ & $.188^{\pm .007}$ & $.163^{\pm .007}$ & $.153^{\pm .007}$ & $.153^{\pm .005}$ & $.154^{\pm .003}$ \\
\midrule
OML & $.105^{\pm .000}$ & $.107^{\pm .000}$ & $.108^{\pm .000}$ & $.110^{\pm .000}$ & $.110^{\pm .000}$ & $.111^{\pm .000}$ \\
OML-Rep & $.104^{\pm .000}$ & $.106^{\pm .000}$ & $.107^{\pm .000}$ & $.108^{\pm .000}$ & $.108^{\pm .000}$ & $.109^{\pm .000}$ \\
TF & $.097^{\pm .000}$ & $.183^{\pm .018}$ & $.208^{\pm .031}$ & $.287^{\pm .053}$ & $.389^{\pm .062}$ & $.347^{\pm .060}$ \\
Linear TF & $.101^{\pm .000}$ & $.125^{\pm .002}$ & $.127^{\pm .002}$ & $.128^{\pm .001}$ & $.132^{\pm .002}$ & $.132^{\pm .001}$ \\
SB-MCL & $.100^{\pm .001}$ & $.103^{\pm .001}$ & $.106^{\pm .001}$ & $.107^{\pm .002}$ & $.108^{\pm .002}$ & $.109^{\pm .002}$ \\
SB-MCL (MAP) & $.100^{\pm .001}$ & $.103^{\pm .001}$ & $.106^{\pm .001}$ & $.107^{\pm .002}$ & $.108^{\pm .002}$ & $.109^{\pm .002}$ \\
\bottomrule
\end{tabular}
\end{table}

\begin{table}[h]
\centering
\caption{CASIA completion with more shots.}
\vskip 0.1in
\small
\begin{tabular}{lrrrrrrr}
\toprule

\multirow{2}{*}{Method} & \multicolumn{5}{c}{Shots} \\
 & 10 & 20 & 50 & 100 & 200 \\
\midrule
Offline & $.146^{\pm .009}$ & $.144^{\pm .006}$ & $.134^{\pm .005}$ & $.144^{\pm .007}$ & $.129^{\pm .007}$ \\
Online & $.290^{\pm .023}$ & $.204^{\pm .008}$ & $.151^{\pm .008}$ & $.152^{\pm .008}$ & $.156^{\pm .008}$ \\
\midrule
OML & $.105^{\pm .000}$ & $.105^{\pm .000}$ & $.105^{\pm .000}$ & $.106^{\pm .000}$ & $.106^{\pm .000}$ \\
OML-Rep & $.104^{\pm .000}$ & $.104^{\pm .000}$ & $.105^{\pm .000}$ & $.106^{\pm .000}$ & $.107^{\pm .000}$ \\
TF & $.097^{\pm .000}$ & $.184^{\pm .019}$ & $.212^{\pm .032}$ & $.301^{\pm .064}$ & $.403^{\pm .062}$ \\
Linear TF & $.101^{\pm .000}$ & $.123^{\pm .002}$ & $.125^{\pm .002}$ & $.126^{\pm .002}$ & $.130^{\pm .002}$ \\
SB-MCL & $.100^{\pm .001}$ & $.100^{\pm .001}$ & $.100^{\pm .001}$ & $.100^{\pm .002}$ & $.100^{\pm .002}$ \\
SB-MCL (MAP) & $.100^{\pm .001}$ & $.100^{\pm .001}$ & $.100^{\pm .001}$ & $.100^{\pm .002}$ & $.100^{\pm .002}$ \\
\bottomrule
\end{tabular}
\end{table}

\begin{table}[h]
\centering
\caption{CASIA rotation with more tasks.}
\vskip 0.1in
\small
\begin{tabular}{lrrrrrrr}
\toprule

\multirow{2}{*}{Method} & \multicolumn{6}{c}{Tasks} \\
 & 10 & 20 & 50 & 100 & 200 & 500 \\
\midrule
Offline & $.544^{\pm .045}$ & $.591^{\pm .047}$ & $.603^{\pm .057}$ & $.510^{\pm .046}$ & $.463^{\pm .044}$ & $.312^{\pm .039}$ \\
Online & $1.079^{\pm .081}$ & $.986^{\pm .073}$ & $.862^{\pm .085}$ & $.616^{\pm .040}$ & $.810^{\pm .059}$ & $.784^{\pm .029}$ \\
\midrule
OML & $.052^{\pm .002}$ & $.052^{\pm .001}$ & $.052^{\pm .001}$ & $.053^{\pm .000}$ & $.053^{\pm .000}$ & $.053^{\pm .001}$ \\
OML-Rep & $.050^{\pm .002}$ & $.050^{\pm .001}$ & $.052^{\pm .001}$ & $.053^{\pm .001}$ & $.055^{\pm .001}$ & $.056^{\pm .001}$ \\
TF & $.034^{\pm .001}$ & $.077^{\pm .003}$ & $.118^{\pm .012}$ & $.122^{\pm .010}$ & $.133^{\pm .006}$ & $.150^{\pm .013}$ \\
Linear TF & $.068^{\pm .002}$ & $.078^{\pm .004}$ & $.086^{\pm .003}$ & $.087^{\pm .002}$ & $.094^{\pm .005}$ & $.091^{\pm .004}$ \\
SB-MCL & $.039^{\pm .001}$ & $.042^{\pm .000}$ & $.045^{\pm .001}$ & $.046^{\pm .000}$ & $.047^{\pm .000}$ & $.047^{\pm .001}$ \\
SB-MCL (MAP) & $.040^{\pm .001}$ & $.042^{\pm .001}$ & $.045^{\pm .001}$ & $.046^{\pm .000}$ & $.047^{\pm .000}$ & $.047^{\pm .000}$ \\
\bottomrule
\end{tabular}
\end{table}

\begin{table}[h]
\centering
\caption{CASIA rotation with more shots.}
\vskip 0.1in
\small
\begin{tabular}{lrrrrrrr}
\toprule

\multirow{2}{*}{Method} & \multicolumn{5}{c}{Shots} \\
 & 10 & 20 & 50 & 100 & 200 \\
\midrule
Offline & $.544^{\pm .045}$ & $.527^{\pm .043}$ & $.465^{\pm .054}$ & $.365^{\pm .053}$ & $.313^{\pm .040}$ \\
Online & $1.079^{\pm .081}$ & $.852^{\pm .062}$ & $.916^{\pm .078}$ & $.649^{\pm .062}$ & $.668^{\pm .073}$ \\
\midrule
OML & $.052^{\pm .002}$ & $.051^{\pm .001}$ & $.052^{\pm .003}$ & $.052^{\pm .002}$ & $.050^{\pm .001}$ \\
OML-Rep & $.050^{\pm .002}$ & $.050^{\pm .001}$ & $.047^{\pm .001}$ & $.046^{\pm .001}$ & $.045^{\pm .000}$ \\
TF & $.034^{\pm .001}$ & $.068^{\pm .004}$ & $.087^{\pm .010}$ & $.086^{\pm .007}$ & $.093^{\pm .008}$ \\
Linear TF & $.068^{\pm .002}$ & $.073^{\pm .004}$ & $.072^{\pm .003}$ & $.075^{\pm .002}$ & $.079^{\pm .006}$ \\
SB-MCL & $.039^{\pm .001}$ & $.038^{\pm .001}$ & $.036^{\pm .001}$ & $.035^{\pm .001}$ & $.035^{\pm .001}$ \\
SB-MCL (MAP) & $.040^{\pm .001}$ & $.039^{\pm .001}$ & $.036^{\pm .001}$ & $.036^{\pm .001}$ & $.035^{\pm .001}$ \\
\bottomrule
\end{tabular}
\end{table}

\begin{table}[h]
\centering
\caption{CASIA VAE with more tasks.}
\vskip 0.1in
\small
\begin{tabular}{lrrrrrrr}
\toprule

\multirow{2}{*}{Method} & \multicolumn{6}{c}{Tasks} \\
 & 10 & 20 & 50 & 100 & 200 & 500 \\
\midrule
Offline & $.664^{\pm .018}$ & $.645^{\pm .027}$ & $.590^{\pm .014}$ & $.571^{\pm .012}$ & $.594^{\pm .017}$ & $.594^{\pm .012}$ \\
Online & $.862^{\pm .009}$ & $.801^{\pm .013}$ & $.760^{\pm .013}$ & $.775^{\pm .019}$ & $.745^{\pm .007}$ & $.736^{\pm .007}$ \\
\midrule
OML & $.442^{\pm .003}$ & $.441^{\pm .003}$ & $.440^{\pm .003}$ & $.440^{\pm .003}$ & $.440^{\pm .003}$ & $.439^{\pm .003}$ \\
OML-Rep & $.454^{\pm .000}$ & $.455^{\pm .001}$ & $.457^{\pm .001}$ & $.457^{\pm .001}$ & $.458^{\pm .001}$ & $.459^{\pm .001}$ \\
SB-MCL & $.428^{\pm .001}$ & $.428^{\pm .001}$ & $.429^{\pm .001}$ & $.429^{\pm .001}$ & $.429^{\pm .001}$ & $.429^{\pm .001}$ \\
SB-MCL (MAP) & $.428^{\pm .001}$ & $.428^{\pm .001}$ & $.429^{\pm .001}$ & $.429^{\pm .001}$ & $.429^{\pm .001}$ & $.429^{\pm .001}$ \\
\bottomrule
\end{tabular}
\end{table}

\begin{table}[h]
\centering
\caption{CASIA VAE with more shots.}
\vskip 0.1in
\small
\begin{tabular}{lrrrrrrr}
\toprule

\multirow{2}{*}{Method} & \multicolumn{5}{c}{Shots} \\
 & 10 & 20 & 50 & 100 & 200 \\
\midrule
Offline & $.664^{\pm .018}$ & $.580^{\pm .014}$ & $.570^{\pm .018}$ & $.564^{\pm .015}$ & $.531^{\pm .014}$ \\
Online & $.862^{\pm .009}$ & $.805^{\pm .016}$ & $.740^{\pm .027}$ & $.780^{\pm .017}$ & $.726^{\pm .017}$ \\
\midrule
OML & $.442^{\pm .003}$ & $.440^{\pm .003}$ & $.440^{\pm .003}$ & $.440^{\pm .002}$ & $.440^{\pm .003}$ \\
OML-Rep & $.454^{\pm .000}$ & $.455^{\pm .002}$ & $.455^{\pm .002}$ & $.456^{\pm .001}$ & $.459^{\pm .001}$ \\
SB-MCL & $.428^{\pm .001}$ & $.428^{\pm .001}$ & $.427^{\pm .001}$ & $.428^{\pm .000}$ & $.428^{\pm .002}$ \\
SB-MCL (MAP) & $.428^{\pm .001}$ & $.427^{\pm .001}$ & $.428^{\pm .001}$ & $.428^{\pm .001}$ & $.428^{\pm .001}$ \\
\bottomrule
\end{tabular}
\end{table}

\begin{table}[h]
\centering
\caption{CASIA DDPM with more tasks.}
\vskip 0.1in
\small
\begin{tabular}{lrrrrrrr}
\toprule

\multirow{2}{*}{Method} & \multicolumn{6}{c}{Tasks} \\
 & 10 & 20 & 50 & 100 & 200 & 500 \\
\midrule
Offline & $.0451^{\pm .0022}$ & $.0408^{\pm .0013}$ & $.0372^{\pm .0017}$ & $.0383^{\pm .0013}$ & $.0382^{\pm .0010}$ & $.0379^{\pm .0008}$ \\
Online & $.1408^{\pm .0032}$ & $.1090^{\pm .0020}$ & $.0787^{\pm .0019}$ & $.0698^{\pm .0016}$ & $.0601^{\pm .0007}$ & $.0511^{\pm .0004}$ \\
\midrule
OML & $.0353^{\pm .0001}$ & $.0352^{\pm .0001}$ & $.0353^{\pm .0001}$ & $.0353^{\pm .0001}$ & $.0353^{\pm .0001}$ & $.0353^{\pm .0001}$ \\
OML-Rep & $.0353^{\pm .0001}$ & $.0353^{\pm .0001}$ & $.0353^{\pm .0001}$ & $.0353^{\pm .0001}$ & $.0352^{\pm .0001}$ & $.0352^{\pm .0001}$ \\
SB-MCL & $.0345^{\pm .0001}$ & $.0347^{\pm .0001}$ & $.0349^{\pm .0001}$ & $.0351^{\pm .0001}$ & $.0351^{\pm .0001}$ & $.0352^{\pm .0000}$ \\
SB-MCL (MAP) & $.0345^{\pm .0001}$ & $.0348^{\pm .0000}$ & $.0350^{\pm .0001}$ & $.0351^{\pm .0001}$ & $.0352^{\pm .0000}$ & $.0353^{\pm .0001}$ \\
\bottomrule
\end{tabular}
\end{table}

\begin{table}[h]
\centering
\caption{CASIA DDPM with more shots.}
\vskip 0.1in
\small
\begin{tabular}{lrrrrrrr}
\toprule

\multirow{2}{*}{Method} & \multicolumn{5}{c}{Shots} \\
 & 10 & 20 & 50 & 100 & 200 \\
\midrule
Offline & $.0451^{\pm .0022}$ & $.0412^{\pm .0011}$ & $.0358^{\pm .0014}$ & $.0380^{\pm .0009}$ & $.0372^{\pm .0018}$ \\
Online & $.1408^{\pm .0032}$ & $.1072^{\pm .0026}$ & $.0826^{\pm .0029}$ & $.0688^{\pm .0020}$ & $.0590^{\pm .0016}$ \\
\midrule
OML & $.0353^{\pm .0002}$ & $.0352^{\pm .0002}$ & $.0353^{\pm .0003}$ & $.0352^{\pm .0001}$ & $.0351^{\pm .0002}$ \\
OML-Rep & $.0353^{\pm .0001}$ & $.0353^{\pm .0002}$ & $.0352^{\pm .0001}$ & $.0353^{\pm .0002}$ & $.0352^{\pm .0001}$ \\
SB-MCL & $.0345^{\pm .0001}$ & $.0345^{\pm .0001}$ & $.0345^{\pm .0001}$ & $.0345^{\pm .0002}$ & $.0345^{\pm .0001}$ \\
SB-MCL (MAP) & $.0345^{\pm .0001}$ & $.0345^{\pm .0001}$ & $.0345^{\pm .0000}$ & $.0344^{\pm .0001}$ & $.0346^{\pm .0001}$ \\
\bottomrule
\end{tabular}
\end{table}

\end{document}